\documentclass[sigconf]{acmart}
\usepackage{colortbl}

\AtBeginDocument{%
  \providecommand\BibTeX{{%
    \normalfont B\kern-0.5em{\scshape i\kern-0.25em b}\kern-0.8em\TeX}}}

\setcopyright{acmlicensed}
\copyrightyear{2024}
\acmYear{2024}
\acmDOI{10.1145/3664647.3680678}

\acmConference[MM'24] {Proceedings of the 32nd ACM International Conference on Multimedia}{October 28--November 1, 2024}{Melbourne, VIC, Australia.}
\acmBooktitle{Proceedings of the 32nd ACM International Conference on Multimedia (MM '24), October 28--November 1, 2024, Melbourne, VIC, Australia}
\acmISBN{979-8-4007-0686-8/24/10}

\begin{document}

\title{Domain-Conditioned Transformer for Fully Test-time Adaptation}


% \author{Yushun Tang\textsuperscript{1}, Shuoshuo Chen\textsuperscript{1}, Jiyuan Jia\textsuperscript{1}, Yi Zhang\textsuperscript{1}, Zhihai He\textsuperscript{1,2}}
% \affiliation{%
%   \institution{\textsuperscript{1}Southern University of Science and Technology, Shenzhen, China \textsuperscript{2}Pengcheng Laboratory, Shenzhen, China}
%   % \streetaddress{1 Th{\o}rv{\"a}ld Circle}
%   % \city{Shenzhen}
%   % \country{China}
%   }
% \email{{tangys2022, chenss2021, jiajy2018, zhangyi2021}@mail.sustech.edu.cn, hezh@sustech.edu.cn}

\author{Yushun Tang}
\affiliation{%
  \institution{Southern University of Science and Technology}
  % \streetaddress{1 Th{\o}rv{\"a}ld Circle}
  \city{Shenzhen}
  \country{China}}
\email{tangys2022@mail.sustech.edu.cn}
\orcid{0000-0002-8350-7637}

\author{Shuoshuo Chen}
\affiliation{%
  \institution{Southern University of Science and Technology}
  % \streetaddress{1 Th{\o}rv{\"a}ld Circle}
  \city{Shenzhen}
  \country{China}}
\email{chenss2021@mail.sustech.edu.cn}
\orcid{0000-0001-5689-5788}

\author{Jiyuan Jia}
\affiliation{%
  \institution{Southern University of Science and Technology}
  % \streetaddress{1 Th{\o}rv{\"a}ld Circle}
  \city{Shenzhen}
  \country{China}}
\email{jiajy2018@mail.sustech.edu.cn}
\orcid{0009-0008-0519-6280}

\author{Yi Zhang}
\affiliation{%
  \institution{Southern University of Science and Technology}
  % \streetaddress{1 Th{\o}rv{\"a}ld Circle}
  \city{Shenzhen}
  \country{China}}
\email{zhangyi2021@mail.sustech.edu.cn}
\orcid{0000-0002-5831-0170}

\author{Zhihai He}
\authornote{Corresponding author.}
\affiliation{%
  \institution{Southern University of Science and Technology, Pengcheng Laboratory}
  % \streetaddress{1 Th{\o}rv{\"a}ld Circle}
  \city{Shenzhen}
  \country{China}}
\email{hezh@sustech.edu.cn}
\orcid{0000-0002-2647-8286}
\renewcommand{\shortauthors}{Yushun Tang, Shuoshuo Chen, Jiyuan Jia, Yi Zhang, and Zhihai He}

\begin{abstract}
  Fully test-time adaptation aims to adapt a network model online based on sequential analysis of input samples during the inference stage. We observe that, when applying a transformer network model into a new domain, the self-attention profiles of image samples in the target domain deviate significantly from those in the source domain, which results in large performance degradation during domain changes. To address this important issue, we propose a new structure for the self-attention modules in the transformer. Specifically, we incorporate three domain-conditioning vectors, called domain conditioners, into the query, key, and value components of the self-attention module.  We learn a network to generate these three domain conditioners from the class token at each transformer network layer. We find that, during fully online test-time adaptation, these domain conditioners at each transform network layer are able to gradually remove the impact of domain shift and largely recover the original self-attention profile. Our extensive experimental results demonstrate that the proposed domain-conditioned transformer significantly improves the online fully test-time domain adaptation performance and outperforms existing state-of-the-art methods by large margins. The code is available at \url{https://github.com/yushuntang/DCT}.
\end{abstract}





\maketitle

% \renewcommand{\thefootnote}{\fnsymbol{footnote}}
% \footnotetext[1]{ Corresponding author.} 
% \renewcommand{\thefootnote}{\arabic{footnote}}

\section{Introduction}
\label{sec:intro}

Transformers have achieved remarkable success in numerous machine learning applications. However, their efficacy can be notably compromised when deployed in unfamiliar domains, primarily due to discrepancies in data distributions between the training datasets in the source domain and the evaluation datasets in the target domain ~\cite{quinonero2008dataset,mirza2021robustness}.  Source-free unsupervised domain adaptation (UDA) ~\cite{liang2020we,Wang_2022_CVPR,li2020model,tang2023cross} aims to recalibrate network models in the absence of any source-domain data samples. Nevertheless, these approaches require complete access to the entire target dataset and retraining of the source model for multiple epochs, making them impractical for real-world applications.
Recently developed test-time adaptation (TTA) methods exhibit promising capabilities in adapting pre-trained models to unlabeled data during testing \cite{Sunttt,wang2020tent,mirza2022norm,tang2023neuro,liang2023comprehensive,yuan2023robust,tang2024learning,chen2024learning,kan2023self}. 
% These strategies can be broadly classified into two primary categories based on their reliance on source domain data: (1) test-time training (TTT) \cite{Sunttt,liu2021ttt++,gandelsman2022test} and (2) fully test-time adaptation \cite{wang2020tent,mirza2022norm,niutowards,tang2023neuro}.
In this work, we focus on the fully test-time adaptation.
Various innovative techniques have been developed to enhance model performance on target domain data online \cite{wang2020tent,zhang2022memo,jing2022variational}. 
% The TENT method \cite{wang2020tent} updates the batch normalization module by minimizing entropy loss. The MEMO method \cite{zhang2022memo} optimizes the entropy of averaged predictions over multiple random augmentations of input samples. Meanwhile, the VMP method \cite{jing2022variational} introduces perturbations into model parameters based on variational Bayesian inference.
These methods, however, are predicated on the availability of a substantial number of samples per mini-batch and a balanced distribution of labels within each mini-batch from the target domain. Such conditions are not always met in practical scenarios.
To counter this limitation, the SAR method \cite{niutowards} introduces a sharpness-aware optimization framework that filters out samples associated with high-gradient noise and promotes a convergence towards a stable, flat minimum in the weight space. 
The RoTTA method \cite{yuan2023robust} introduces category-balanced sampling, aiming to enhance the robustness of the model against label imbalances.

Recently, transformer-based methods have achieved remarkable success in various machine learning tasks due to their powerful self-attention capabilities. 
In this work, we propose to explore how transformer networks can be successfully adapted to new domains during the testing stage.
During our experiments, we find that when transformer models are applied to new domains, their self-attention distance profiles, defined as the spatial distribution of self-attention between tokens, for image samples in the target domain deviate significantly from those in the source domain.
Note that the self-attention is one of the core modules in the transformer network design. 
Once this self-attention profile has been perturbed by the domain changes or corruptions, the performance of the transformer model will degrade significantly. 
A research question arises: \textit{how do we remove these perturbations caused by the domain shifts from the self-attention profile so as to improve the test-time adaptation performance of transformer models? }

\begin{figure}[!t]
    \centering
    \includegraphics[width = \linewidth]{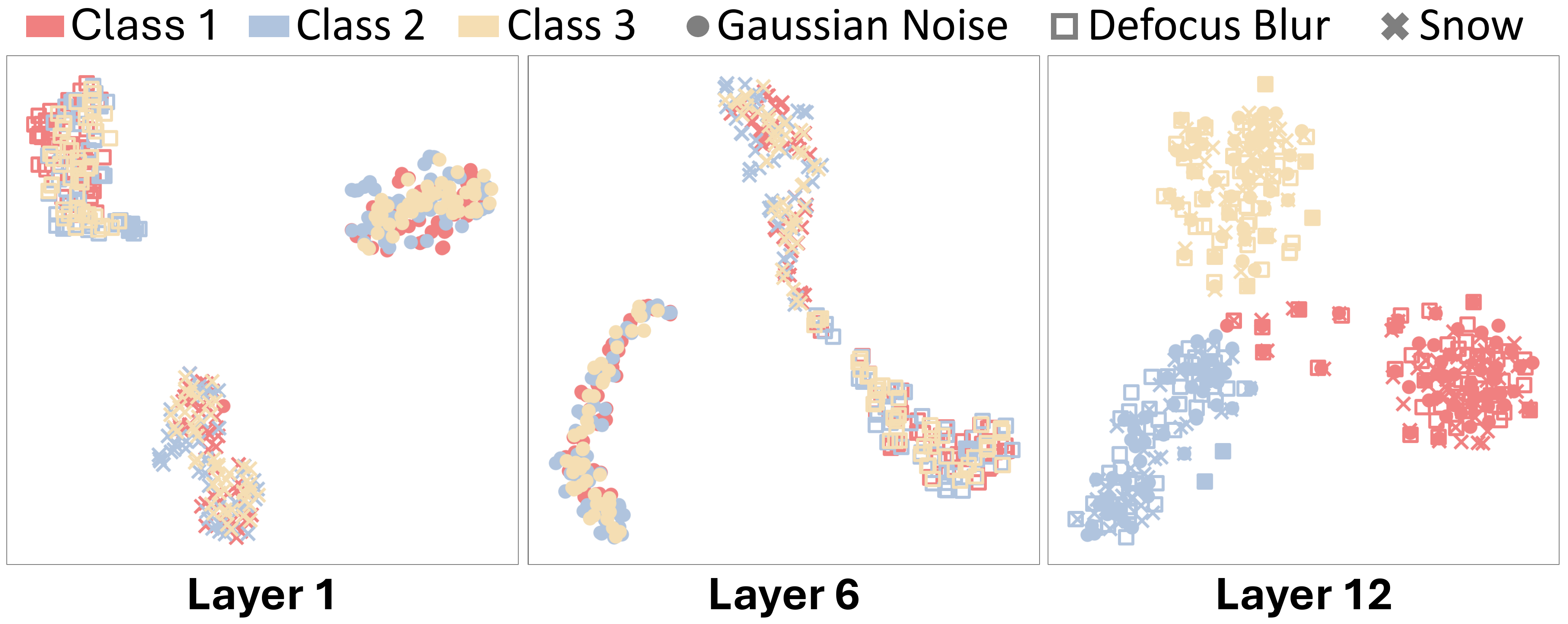}
    \caption{Visualization of output class tokens across various layers of our adapted ViT-B/16 network in ImageNet-C dataset. Different corruptions represent different domains. In layer 1, the features exhibit domain-separability and class-inseparability due to the presence of domain shift, with a considerable distance between domains and a small distance between classes. Our DCT method effectively mitigates the influence of domain shift over successive layers. Consequently, the domain distance decreases while the class distance increases, leading to the features domain-inseparable yet class-separable across the layers of the Domain-Conditioned Transformer.}
    \label{fig: feature_layer_change_class}
% \vspace{-20pt}
\end{figure}

To address this challenge, we propose to 
explore a new approach, called \textit{domain-conditioned transformer}, for fully test-time adaptation of transformer models.
Specifically, we introduce three domain-conditioning vectors, called \textit{domain conditioners}, into the query, key, and value components of the self-attention modules. These domain conditioners are designed to capture domain-specific perturbation information and remove these perturbations layer by layer. 
We learn a domain conditioner generation network to generate these domain conditioners from class tokens of the previous network layers, containing both semantic and domain information. 
% We observe that, based on this new self-attention module equipped with these domain conditioners, the self-attention distance profiles of image samples can be largely restored. 
We observe that, our proposed approach gradually mitigates the impact of domain shift. As shown in Figure \ref{fig: feature_layer_change_class}, the domain information is gradually removed and the class information is enhanced across the domain-conditioned transformer layers. This gradual adjustment process facilitates the recovery of the original self-attention profile, allowing the model to maintain its performance across diverse domains. 
Our extensive experimental results demonstrate that the proposed domain-conditioned transformer significantly improves the online test-time domain adaptation performance and outperforms existing state-of-the-art methods by large margins.

\section{Related Work}
\label{sec:related_work}
This work is related to test-time adaptation, source-free unsupervised domain adaptation, parameter efficient transfer learning, and prompt learning. 

\subsection{Test-time Adaptation}
Test-time adaptation (TTA) is a strategy designed to refine a pre-trained model so that it can better accommodate unlabeled data that may exhibit shifts in domain characteristics during the inference phase. There are two major approaches: \textit{test-time training} \cite{Sunttt,liu2021ttt++,gandelsman2022test} and \textit{fully test-time adaptation} \cite{wang2020tent,mirza2022norm,niutowards}. 
The pioneering work in the domain of test-time training (TTT) was introduced by Sun et al., where the parameters of the feature extraction network are dynamically refined during inference using a self-supervised loss function applied to a surrogate learning task \cite{Sunttt}. It is important to recognize that these TTT methodologies necessitate a period of specialized training within the source domain to ensure the model's adaptability.

In contrast, fully test-time adaptation methods adjust pre-trained models in real-time during inference, bypassing the need for access to source domain data. The TENT approach \cite{wang2020tent} pioneers this concept by calibrating the Batch Normalization (BN) layers, allowing the model to adapt to new data distributions. Similarly, the NHL method \cite{tang2023neuro} introduced unsupervised learning of early-layer features to TTA, inspired by the principles of Hebbian learning from neurobiological studies.
Innovative strategies have also emerged that focus on transforming model inputs rather than the underlying network parameters. The DDA method \cite{gao2022back}, for instance, employs a diffusion model to map target domain inputs into the source domain space during the testing phase. Another approach modifies the inputs at the image level by learning visual prompts, with the underlying model parameters remaining constant, as proposed in \cite{gan2023decorate}.
It has been noted that existing online model updating methods suffer from performance degradations due to sample imbalances and distribution shifts. To address these challenges, the SAR method \cite{niutowards} introduces a robust optimization strategy that filters out samples causing high-gradient noise and encourages the model weights to converge towards a stable minimum. DELTA \cite{zhao2023delta}  uses moving averaged statistics to perform the online adaptation of the normalized features.

\subsection{Parameter Efficient Transfer Learning}
As the model size grows rapidly with the development of foundation models, there has been a growing interest among researchers focusing on Parameter Efficient Transfer Learning \cite{DBLP:conf/emnlp/LesterAC21,houlsby2019parameter}. This area of study focuses on adapting large-scale pre-trained models to different downstream tasks with minimal modification of parameters. 
These methods strategically select a subset of pre-trained parameters and, if needed, introduce a limited number of additional parameters into a pre-trained network. These selected parameters are updated specifically for new tasks, while the majority of the original model parameters are frozen to ensure efficiency and effectiveness.
For instance, the method proposed by \cite{liu2022few} introduces learnable vectors to rescale keys, values in attention mechanisms, and inner activations in position-wise feed-forward networks through element-wise multiplication. \textit{Diff pruning} \cite{guo2021parameter} learns an adaptively pruned task-specific “diff” vector extending the original pre-trained parameters. BitFit \cite{zaken2022bitfit} employs sparse fine-tuning, where only the bias terms (or a subset of them) are modified. AdaptFormer \cite{chen2022adaptformer} adapts pre-trained Vision Transformers (ViTs) for various vision tasks by replacing the original MLP block with a trainable down-up bottleneck module. In contrast, LST \cite{sung2022lst} introduces a separate ladder side network, a smaller network that uses intermediate activations as inputs based on shortcut connections, rather than inserting additional parameters inside the backbone networks.
It should be noted that all of these parameter efficient transfer learning methods adapt pre-trained models to downstream tasks through supervised learning. In this work, we perform a parameter efficient online adaptation in an unsupervised manner.

% Pro-tuning: Unified Prompt Tuning for Vision Tasks

\subsection{Prompt Learning}
Prompt Learning is a concept initially from Natural Language Processing (NLP), involving adding prompts to input text to guide pre-trained language models in specific downstream tasks \cite{liu2023promptsurvey}. 
% Innovations in recent methodologies \cite{DBLP:conf/emnlp/LesterAC21, liu2021ptuningv2, li2021prefix} have led to the representation of prompts as task-specific vectors, trainable through error back-propagation. This approach to prompt tuning allows for learning prompts directly from task data within the input embedding space, thus requiring minimal parameter updates during adaptation.
The application of prompt learning has extended to vision-language models with notable success \cite{zhou2022coop, zhou2022conditional, shu2022test, zhang2024cross,zhang2024concept,zhang2023bdc}. For instance, CoOp \cite{zhou2022coop} adjusts the prompt of the text encoder in CLIP \cite{radford2021clip} to better align with downstream tasks. CoCoOp \cite{zhou2022conditional} further refines this by conditioning the learned prompt on the model's input data, tackling issues related to out-of-distribution samples. TPT \cite{shu2022test} enhances the text encoder's prompt of CLIP during the test phase, aiming to improve generalization through entropy minimization and confidence-based selection.
Beyond the textual domain, techniques have emerged for learning visual prompts in computer vision tasks. The method proposed by \cite{gan2023decorate} is centered on the continuous application of Test-Time Adaptation, where both domain-specific and domain-agnostic visual prompts, are dynamically learned and integrated into target images at pixel level. 
Visual Prompt Tuning (VPT) \cite{jia2022visual, gao2022visual} represents another advancement in this field. It incorporates task-specific, learnable parameters directly into the input sequence at each layer of the Vision Transformer (ViT) encoder. Notably, this is achieved while keeping the pre-trained transformer encoder's backbone frozen during subsequent training phases.
% This approach has found applications in transfer learning for image synthesis \cite{sohn2022visualgan}. 
However, it's important to acknowledge that the VPT method introduces a considerable number of new tokens into the transformer. This addition can significantly amplify the computational cost of the self-attention mechanism, given that the complexity of this mechanism is proportional to the square of the number of input tokens. In this work, we generate only one extra domain-conditioner token with a newly introduced generator conditioned by the class token in each transformer layer.

\subsection{Unique Contributions}
In this work, we propose to explore fully online test-time adaptation of transformer models by designing adaptable self-attention modules. 
Compared to existing work, the {major contributions} of this work can be summarized as follows: (1) We introduce a new design of the self-attention module in the transformer networks, which is able to capture the domain-specific characteristics of test samples in the target domain and is able to clean up the domain shift perturbations in the test samples.
(2) We learn a lightweight neural network called domain-conditioner generator to generate the domain conditioners from the class token at each layer, enabling the transformer model to better align its self-attention profiles with the source domain. 
(3) Our experimental results demonstrate that our proposed domain-conditioned transformer is able to significantly improve the online domain adaptation performance of transformer models, outperforming the state-of-the-art method in fully test-time domain adaptation across multiple popular benchmark datasets and test conditions.

\section{Method}
\label{sec:methods}

In this section, we present our method of Domain-Conditioned Transformer (DCT) for fully test-time adaptation.

\subsection{Method Overview}
Suppose we have a model $\mathcal{M} = f_{\theta_s}(y|X_s)$ with parameters $\theta_s$, successfully trained on source data $\{X_s\}$ with corresponding labels $\{Y_s\}$. During fully test-time adaptation, we are provided with target data $\{X_t\}$ along with unknown labels $\{Y_t\}$. Our objective is to adapt the trained model online in an unsupervised manner during testing.
In this scenario, we receive a sequence of input sample batches $\{\mathbf{B}_1, \mathbf{B}_2, ..., \mathbf{B}_T\}$ from the target data $\{X_t\}$. It should be noted that, during each adaptation step $t$, the network model can only rely on the $t$-th batch of test samples, denoted as $\mathbf{B}_t$. Following the \textit{wild} test-time adaptation setting outlined in SAR \cite{niutowards}, it's possible that each mini-batch $\mathbf{B}_t$ may contain only one sample, and the samples within the mini-batch can be imbalanced.

\begin{figure}[!t]
% \vspace{-0.6cm}
\setlength{\abovecaptionskip}{-0cm}
    \centering
    \includegraphics[width = \linewidth]{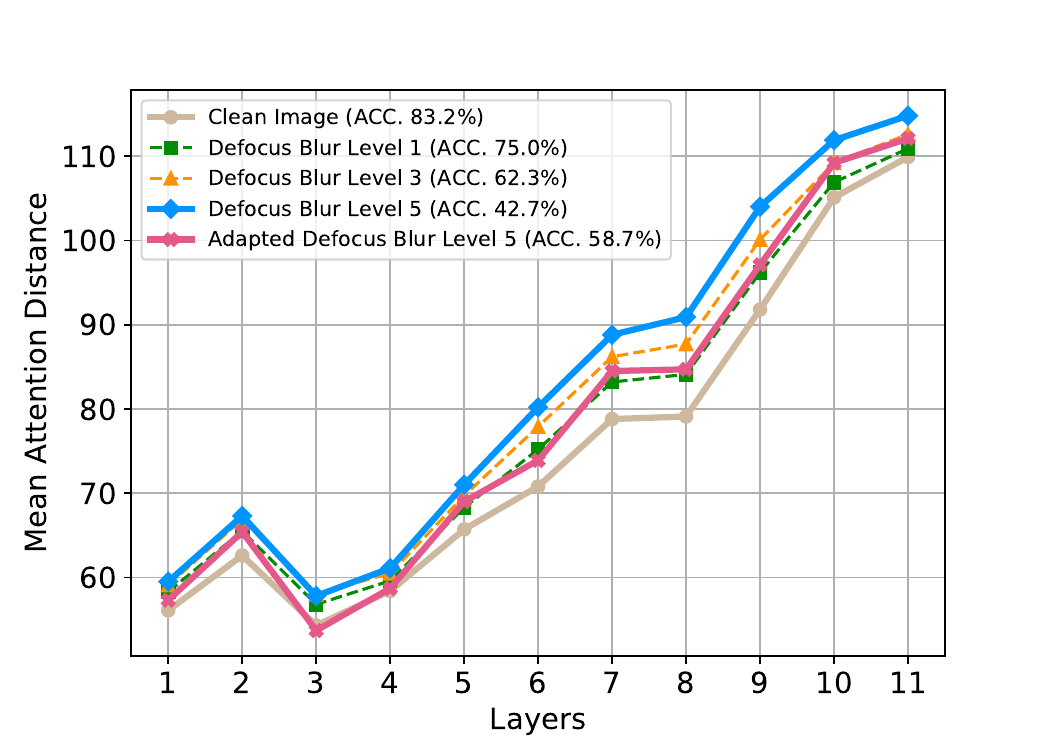}
    \caption{Size of the attended area by transformer network depth. Each dot on the figure represents the mean attention distance calculated across 128 example images, considering all heads at a specific layer.}
    \label{fig:attention_distance}
\end{figure}

\begin{figure*}[!ht]
\setlength{\abovecaptionskip}{0.1cm}
\setlength{\belowcaptionskip}{-0.3cm}
    \centering
    \includegraphics[width = 0.95\textwidth]{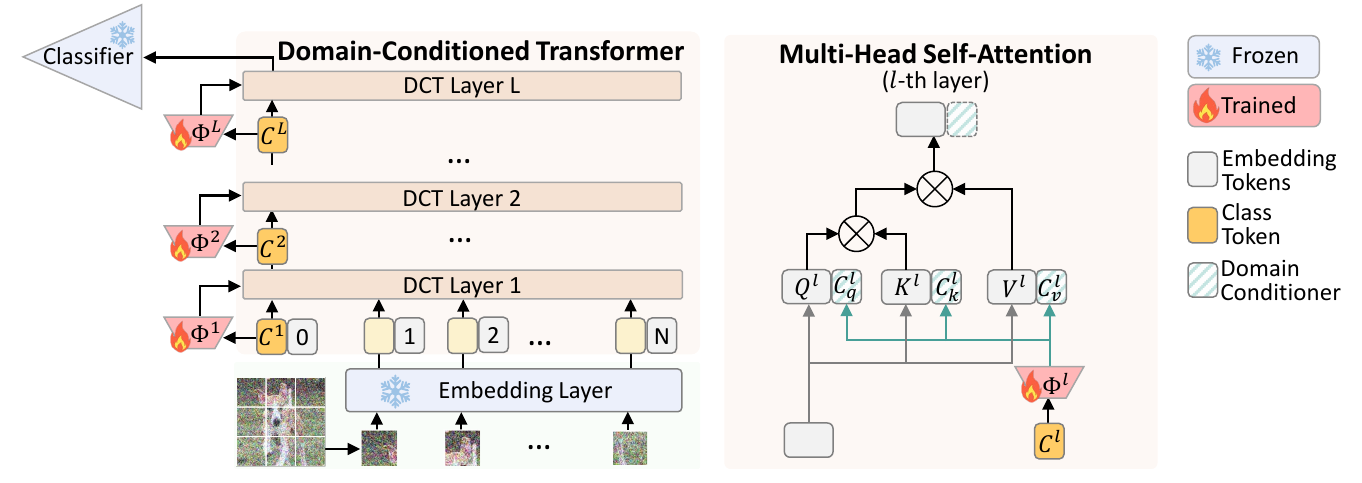}
    \caption{An overview of the proposed DCT method. During inference in the target domain, the domain conditioners generator $\Phi^{l}$ and LN layers are updated before making a prediction given each mini-batch testing sample. The domain-conditioned transformer (\textbf{Left}). The details of the self-attention head in each layer (\textbf{Right}).}
    \label{fig: framework}
% \vspace{-0.5cm}
\end{figure*}

When adapting transformer-based models to new domains, we observe that 
their self-attention distance profiles, defined as the spatial distribution of self-attention between tokens, for image samples in the target domain deviate significantly from those in the source domain. 
Let $\{\Omega_i^l|1\le i\le N \}$ be the set of $N$ embedding tokens at the network layer $l$, and $\mathbb{C}=[c^l_{ij}]_{N\times N}$ be their self-attention weight matrix. 
Let $\mathbb{D}=[d^l_{ij}]_{N\times N}$ be the distance matrix for these tokens where $d^l_{ij}$ represents the pixel distance in the original input image between tokens $i$ and $j$.
The attention distance 
\cite{dosovitskiyimage,raghu2021vision}
for network layer $l$ is then defined as
\begin{equation}
d(l) = \sum_{i,j}c^l_{ij}\cdot d^l_{ij}.
\end{equation}
The attention distance $d^{l}$  for all network layers is referred to as the self-attention profile, denoted as 
$\{d(l)\}$. 
In Figure \ref{fig:attention_distance}, we plot the attention profile for the clean image, images with different levels of domain corruptions (Defocus Blur), and the attention profile recovered by our DCT method. 
Conceptually, this attention distance is similar to the receptive field size in Convolutional Neural Networks (CNNs). It indicates that lower layers of the ViT model tend to focus on local regions more, as evidenced by a lower mean attention distance. In contrast, higher layers primarily integrate global information, leading to a higher attention distance.  When there is a data distribution shift during testing in the target domain, the attention distance distribution is shifted.
As illustrated in Figure \ref{fig:attention_distance}, as the level of image corruption increases, the corresponding attention distance becomes large.  
Once this self-attention profile has been perturbed by the domain changes or corruptions, the performance of the transformer model will degrade significantly. 
It can be seen that, using the DCT method, the attention distance profiles of the corrupted images can be largely recovered to their minimum level (Level 1), approaching the attention distance profile of the clean image. Meanwhile, the Attention Rollout for the target samples exhibits improved focus compared to the source model, as illustrated in Figure \ref{fig:attention}.

In this work, we propose to introduce a new self-attention structure that is able to capture the domain perturbations and gradually remove them from the image features.
As shown in Figure \ref{fig: framework}, at layer $l$ of the proposed domain-conditioned transformer (DCT), we append three domain-conditioning vectors, $[C^l_q, C^l_k, C^l_v]$, into the query, key, and value components $[Q^l, K^l, V^l]$ of its self-attention module. 
At each transformer network layer, we learn a lightweight neural network (domain-conditioner generator $\Phi^l$) to generate these three domain conditioners $[C^l_q, C^l_k, C^l_v]$ from the class token $C^l$. During fully test-time adaptation, 
$\Phi^l$ is updated during the inference process. 
These domain conditioners at each transform network layer are able to gradually remove the impact of domain shift and significantly recover the original self-attention profile.
In the following section, we explain the proposed DCT method in more detail.

\subsection{Domain-Conditioned Self-Attention}
Self-attention mechanisms have demonstrated remarkable performance in various computer vision tasks by capturing correlation between image patches. The output matrix of self-attention is defined as: 
\begin{equation}
    \text{Attention}(Q, K, V) = \text{softmax}\left(\frac{QK^\top}{\sqrt{d}}\right)V,
\end{equation}
where $d$ represents the dimensions of the query, key, and value. For convenience, we consider layer $l$ of the network and omit the superscript $l$ here in this section.
The self-attention weights are computed from the correlation between patch embeddings. Certainly, the self-attention distance profile defined in the previous section changes when the input image is perturbed by domain shifts. 
Write the query $Q \in \mathbb{R}^{n \times d}$, key $K \in \mathbb{R}^{n \times d}$, and value $V \in \mathbb{R}^{n \times d}$ matrices of the self-attention mechanism for the $n$ embedding tokens (n = N+1, including class token) as $Q = [\mathbf{q_{1}}, \mathbf{q_{2}}, \cdots, \mathbf{q_{n}}]^\top$, $K = [\mathbf{k_{1}}, \mathbf{k_{2}}, \cdots, \mathbf{k_{n}}]^\top$, $V = [\mathbf{v_{1}}, \mathbf{v_{2}}, \cdots, \mathbf{v_{n}}]^\top$.
The original correlation matrix 
% $ A \in \mathbb{R}^{n \times n}$ 
between $Q$ and $K$ is denoted by 
$    QK^\top = 
    [\alpha_{i,j}]_{n \times n}.
$
In our proposed DCT method, we introduce three domain conditioning vectors $C_{q}\in \mathbb{R}^{1 \times d}$, $C_{k}\in \mathbb{R}^{1 \times d}$, and $C_{v}\in \mathbb{R}^{1 \times d}$ and append them to the query, key, and value matrices, respectively, and obtain the following augmented 
query $\bar{Q} \in \mathbb{R}^{(n+1) \times d}$, key $\bar{K} \in \mathbb{R}^{(n+1) \times d}$, and value $\bar{V} \in \mathbb{R}^{(n+1) \times d}$:
\begin{equation}
    \bar{Q} = 
    \begin{bmatrix}
        Q \\
        C_{q}
    \end{bmatrix}, \quad
    \bar{K} = 
    \begin{bmatrix}
        K \\
        C_{k}
    \end{bmatrix}, \quad
    \bar{V} = 
    \begin{bmatrix}
        V \\
        C_{v}
    \end{bmatrix}.
\end{equation}
The correlation matrix between $\bar{Q}$ and $\bar{K}$ with domain conditioners is:
\begin{equation}
\small
\begin{aligned}
    \bar{Q} \bar{K}^\top  
    =
    \begin{bmatrix}
        \alpha_{1,1}  & \alpha_{1,2}  & \cdots & \alpha_{1,n}  & \mathbf{q_{1}}C_{k}^\top\\
        \alpha_{2,1}  & \alpha_{2,2}  & \cdots & \vdots & \mathbf{q_{2}}C_{k}^\top\\
        \vdots & \vdots & \ddots & \vdots & \vdots \\
        \alpha_{n,1}  & \alpha_{n,2}  & \cdots & \alpha_{n,n}  & \mathbf{q_{n}}C_{k}^\top \\
        C_{q}\mathbf{k_{1}}^\top & C_{q}\mathbf{k_{2}}^\top & \cdots & C_{q}\mathbf{k_{n}}^\top & C_{q}C_{k}^\top
    \end{bmatrix}
\end{aligned}
\end{equation}    
Now, the new self-attention weight matrix
$W=[w_{i,j}]_{(n+1)\times (n+1)}$ is given by:

\begin{equation}
\label{eq:softmax}
\small
\begin{aligned}
    &w_{i,j} = \text{softmax}\left( \bar{Q} \bar{K}^\top\right) \\&=
    \begin{cases}
        \frac{\exp({\alpha_{i,j}})}{\sum_{j=1}^{n} \exp({\alpha_{i,j}}) + \exp({\mathbf{q_{i}}C_{k}^\top})}, \quad  &i \neq n+1, j \neq n+1;\\
        \frac{e^{\mathbf{q_{i}}C_{k}^\top}}{\sum_{j=1}^{n} \exp({\alpha_{i,j}}) + \exp({\mathbf{q_{i}}C_{k}^\top})},\quad &i \neq n+1, j = n+1;\\        \frac{\exp({C_{q}\mathbf{k_{j}}^\top})}{\sum_{j=1}^{n} \exp({C_{q}\mathbf{k_{j}}^\top}) + \exp({C_{q}C_{k}^\top})},\quad &i = n+1, j \neq n+1;\\       \frac{\exp({C_{q}C_{k}^\top})}{\sum_{j=1}^{n} \exp({C_{q}\mathbf{k_{j}}^\top}) + \exp({C_{q}C_{k}^\top})},\quad &i = n+1, j = n+1.\\
    \end{cases}
\end{aligned}
\end{equation}
The conditioned self-attention output is:
\begin{equation}
\label{eq:conditioned attention}
\begin{aligned}
    &\text{Attention}(\bar{Q}, \bar{K}, \bar{V}) = \text{softmax}\left(\frac{\bar{Q}\bar{K}^\top}{\sqrt{d}}\right)\bar{V}.
\end{aligned}
\end{equation}
From (\ref{eq:softmax}) and (\ref{eq:conditioned attention}), we can see that the original self-attention weights have been modified by the domain conditioner.
The introduction of domain conditioners adds an additional context to the attention mechanism.
The softmax score of $\alpha_{i,j}$  is re-calibrated to consider both input data and the contextual information of the target domain.
In our proposed DCT method for fully test time adaptation, these domain conditioners
$[C_q, C_k, C_v]$ are generated by a network that is learned online during the test-time adaptation process, which will be explained in the following section. 

\subsection{Domain Conditioner Generator Networks}

At each transformer network layer, we introduce a dedicated light-weight network $\Phi_l$, called \textit{ domain conditioner generator}
to generate the three domain conditioners 
$[C^l_q, C^l_k, C^l_v]$ from the class token $C^l$:
\begin{equation}
[C^l_q, C^l_k, C^l_v] = \Phi^l(C^l).
\end{equation}
This domain conditioner generator network is learned during the test-time adaptation process.
Specifically, in the current mini-batch $\mathbf{B}_t$,  
when training the network $\Phi_l$, we aim to 
minimize the loss function $\mathcal{L}(\theta_t; x)$ with respect to the network's trainable parameters $\theta_t$.
In the context of test-time adaptation, the loss function $\mathcal{L}(\theta_t; x)$ is often based on the entropy of the predictions for a given batch of data. However, we've observed that straightforward minimization of this loss can lead to a phenomenon known as model collapse. To address this issue, we employ a combination of two strategies: reliable entropy minimization and sharpness-aware minimization, as discussed in \cite{foret2020sharpness, niutowards}. The reliable entropy minimization technique involves discarding test samples that exhibit high entropy, thereby mitigating the influence of noisy or unreliable data on the model's adaptation process.
On the other hand, sharpness-aware minimization promotes a convergence of the model weights towards a flat minimum to enhance the robustness of the adaptation process. The combined optimization loss can be expressed as follows:
\begin{equation}\label{eq: loss}
     \mathcal{L}(\theta_t; x) = \mathbb{I}[\mathbb{E}(\theta_t; x)<E_0]\cdot \mathbb{E}(\theta_t; \mathbf{B}_t),
\end{equation}
where $\mathbb{E}$ is the entropy loss function, and $\mathbb{I}[\mathbb{E}(\theta_t; x)<E_0]$ is the mask to filter out test samples when the corresponding entropy is larger than the threshold $E_0$.

\begin{figure}[!htbp]
% \vspace{-10pt}
\setlength{\abovecaptionskip}{0.1cm}
    \centering
    \includegraphics[width = 0.95\linewidth]{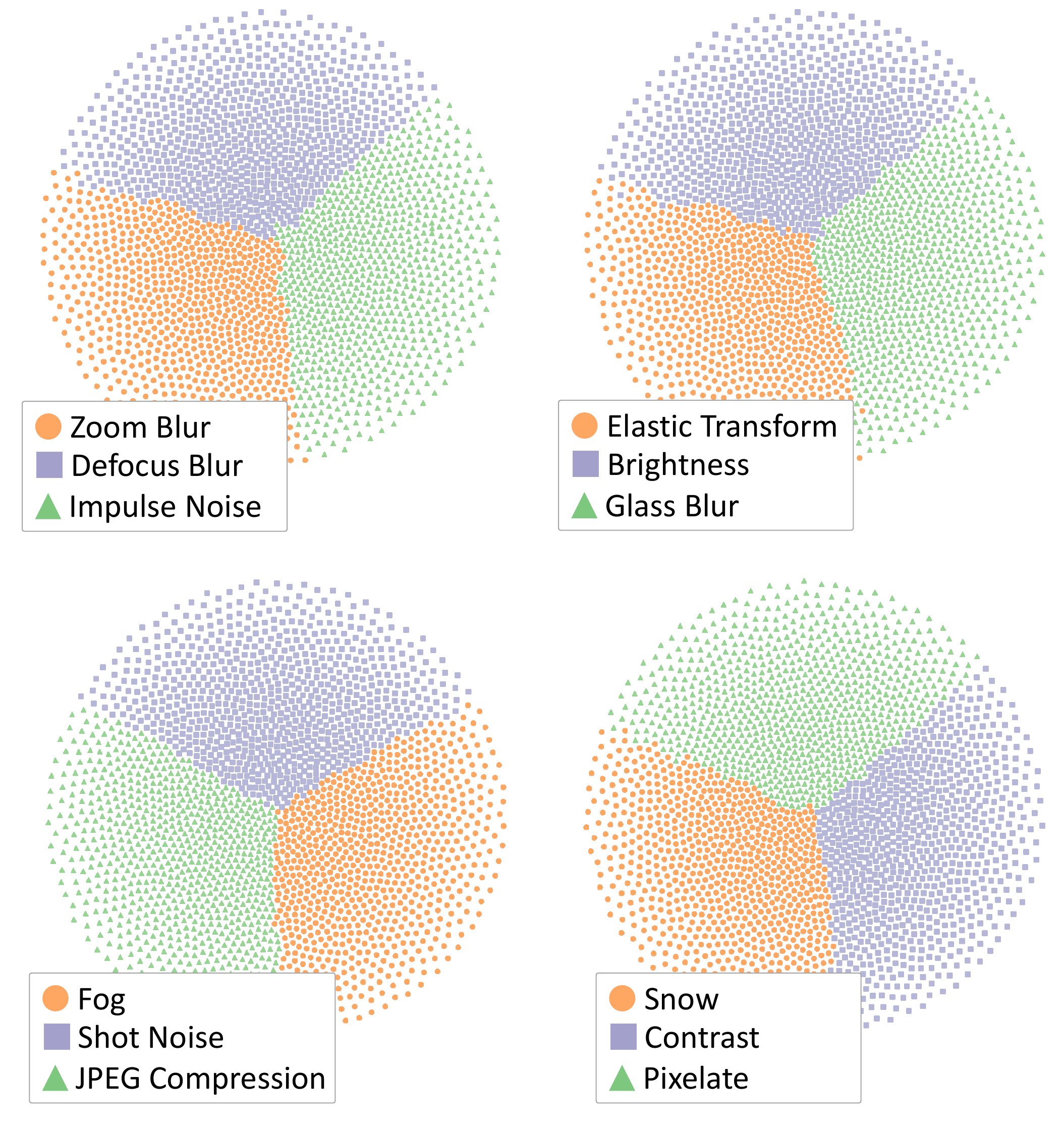}
    \caption{Visualization of the domain conditioners after the entire adaptation process for different domains in ImageNet-C from the first vision transformer layers.}
    \label{fig: qkv}
% \vspace{-10pt}
\end{figure}

Figure \ref{fig: qkv} shows the t-SNE plot visualization of the domain conditioners $[C^l_q, C^l_k, C^l_v]$ in layer 1 for samples from different target domains, with different domain corruptions, plotted with three different colors.   
We can see that, samples from the same domain, although from totally different classes, aggregate together. This suggests that the domain conditioners $[C^l_q, C^l_k, C^l_v]$ generated by the learned network $\Phi^l$ are able to capture the domain characteristics.

\begin{figure*}[!ht]
\setlength{\abovecaptionskip}{0.2cm}
\setlength{\belowcaptionskip}{-0.2cm}
    \centering
    \includegraphics[width = \textwidth]{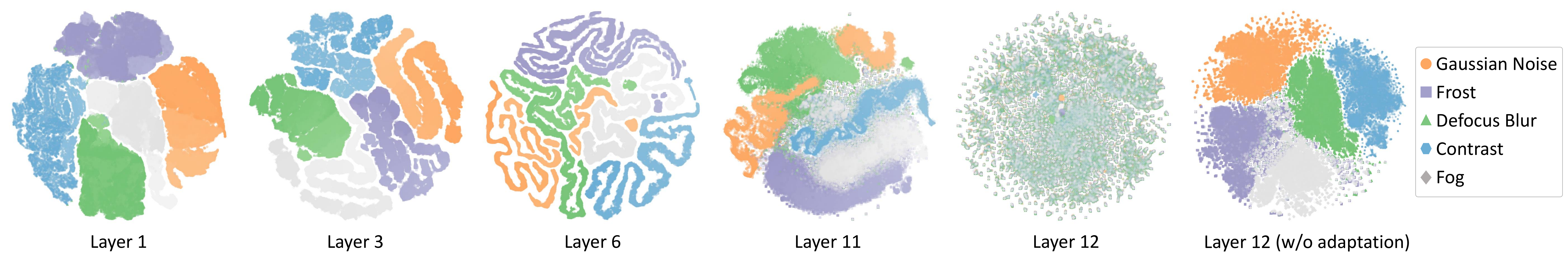}
    \caption{Visualization of output class tokens from different vision transformer layers. The first 5 plots show the features from various layers of our DCT, and the last plot shows the features of the source model for comparison.}
    \label{fig: feature_layer_change}
% \vspace{-0.5cm}
\end{figure*}

We observe that the domain conditioners play a crucial role in capturing and gradually removing the domain perturbation from the image features throughout the transformer network layers. Figure \ref{fig: feature_layer_change} shows the class tokens $C^l$ at different layers of our domain-conditioned transformer. 
Specifically, the first 5 plots show the class token at layers 1, 3, 6, 11, and 12 of the domain-conditioned transformer. Each plot shows the samples from 5 different target domains (Gaussian Noise, Frost, Defocus Blur, Contrast, and Fog) with each domain being plotted with a different color. 
We can see that, with the proposed domain conditioning learning and adaptation, the domain information is being gradually removed from the class tokens. 
In the 5-th plot for layer 12, we can hardly see any domain difference among these samples.
For comparison, in the 6-th plot, we also show the class token of layer 12 from the source model without using the proposed DCT method. 
We can see that the domain information is clearly seen in the final layer of the transformer model. This will significantly degrade the performance of the network in the target domain. 

\begin{table*}[!htbp]
\begin{center}
\caption{Classification Accuracy (\%) for each corruption in \textbf{ImageNet-C} under \textbf{Normal} at the highest severity (Level 5). The best result is shown in \textbf{bold}.}
\label{table: normal}
\resizebox{\linewidth}{!}
{
\begin{tabular}{l|ccccccccccccccc|c}
\toprule
Method & gaus & shot & impul & defcs & gls & mtn & zm & snw & frst & fg & brt & cnt & els & px & jpg & Avg.\\	
\midrule
Source & 46.9 & 47.6 & 46.9 & 42.7 & 34.2 & 50.5 & 44.7 & 56.9 & 52.6 & 56.5 & 76.1 & 31.8 & 46.7 & 65.5 & 66.0 & 51.0 \\
% EATA \cite{niu2022efficient} &  \\
T3A \cite{iwasawa2021test} & 16.6 & 11.8 & 16.4 & 29.9 & 24.3 & 34.5 & 28.5 & 15.9 & 27.0 & 49.1 & 56.1 & 44.8 & 33.3 & 45.1 & 49.4 & 32.2\\
CoTTA \cite{wang2022continual} & 40.3 & 31.8 & 39.6 & 35.5 & 33.1 & 46.9 & 37.3 & 2.9 & 46.4 & 59.1 & 71.7 & 55.5 & 46.4 & 59.4 & 59.0 & 44.4\\
DDA~\cite{gao2022back} &   52.5  & 54.0  & 52.1  & 33.8  & 40.6  & 33.3  & 30.2  & 29.7  & 35.0  & 5.0  & 48.6  & 2.7  & 50.0  & 60.0  & 58.8  & 39.1 \\
MEMO~\cite{zhang2022memo} & 58.1 & 59.1 & 58.5 	& 51.6 & 41.2 & 57.1 & 52.4 & 64.1 & 59.0 & 62.7 & \textbf{80.3} & 44.6 & 52.8 & 72.2 & 72.1 & 59.1 \\
AdaContrast \cite{chen2022contrastive} & 54.4 & 55.8 & 55.8 & 52.5 & 42.2 & 58.7 & 54.3 & 64.6 & 60.1 & 66.4 & 76.8 & 53.7 & 61.7 & 71.9 & 69.6 & 59.9 \\
CFA \cite{kojima2022robustvit} & 56.9 & 58.0 & 58.1 & 54.4 & 48.9 & 59.9 & 56.6 & 66.4 & 64.1 & 67.7 & 79.0 & 58.8 & 64.3 & 71.7 & 70.2 & 62.4 \\
TENT~\cite{wang2020tent} & 57.6 & 58.9 & 58.9 & 57.6 & 54.3 & 61.0 & 57.5 & 65.7 & 54.1 & 69.1 & 78.7 & 62.4 & 62.5 & 72.5 & 70.6 & 62.8 \\
DePT-G \cite{gao2022visual} &53.7 & 55.7 & 55.8 & 58.2 & 56.0 & 61.8 & 57.1 & 69.2 & 66.6 & 72.2 & 76.3 & 63.2 & 67.9 & 71.8 & 68.2 & 63.6\\
SAR~\cite{niutowards} & 58.0 & 59.2 & 59.0 & 58.0 & 54.7 & 61.2 & 57.9 & 66.1 & 64.4 & 68.6 & 78.7 & 62.4 & 62.9 & 72.5 & 70.5 & 63.6 \\ 
\rowcolor{gray!20}
\textbf{Ours} & \textbf{58.8} & \textbf{60.2} & \textbf{60.1} & \textbf{58.7} & \textbf{58.9} & \textbf{63.2} & \textbf{62.9} & \textbf{69.4} & \textbf{68.1} & \textbf{73.2} & 79.6 & \textbf{65.1} & \textbf{69.0} & \textbf{74.4} & \textbf{72.3} & \textbf{66.3} \\ 
\rowcolor{gray!20}
& ${\pm0.2}$ & ${\pm0.1}$ & ${\pm0.1}$ & ${\pm0.1}$ & ${\pm0.3}$ & ${\pm0.1}$ & ${\pm0.2}$ & ${\pm0.2}$ & ${\pm0.3}$ & ${\pm0.0}$ & ${\pm0.1}$ & ${\pm0.2}$ & ${\pm0.1}$ & ${\pm0.1}$ & ${\pm0.2}$ & ${\pm0.0}$ \\ 
\bottomrule
\end{tabular}
}
% \vspace{-10pt}
\end{center}
\end{table*}

\section{Experiments}
\label{sec:exp}

In this section, we conduct experiments on multiple online test-time adaptation settings and multiple dataset benchmarks to evaluate the performance of our proposed DCT method.

\subsection{Benchmark Datasets and Baselines} 
In our experiments, we select the widely used \textbf{ImageNet-C} benchmark dataset \cite{hendrycks2018benchmarking}, consisting of $50,000$ instances distributed across $1,000$ classes. Additionally, we test in \textbf{ImageNet-R} \cite{hendrycks2021many}, a dataset containing 30,000 images presenting diverse artistic renditions of 200 classes from the ImageNet dataset.
% Furthermore, our evaluation of Test-Time Adaptation (TTA) performance encompasses benchmark datasets tailored for domain adaptation and domain generalization tasks. 
The results also include \textbf{VisDA-2021} \cite{bashkirova2022visda}, a dataset designed to assess models' ability to adapt to novel test distributions and effectively handle distributional shifts.
% VisDA-2021 is specifically designed to assess models' ability to adapt to novel test distributions and effectively handle distributional shifts. 
Additionally, we utilize the \textbf{Office-Home} \cite{venkateswara2017deep} dataset, which has a total of $15,500$ images spanning $65$ object categories across four distinct domains.
We compare our proposed DCT method against the following fully online test-time adaptation methods: no adaptation which is the source model, T3A, CoTTA, DDA, MEMO, TENT, AdaContrast, CFA, DePT-G, and SAR.

% (1) \textbf{Source:} the baseline model is trained only on the source data without any fine-tuning during the test process.
% (2) \textbf{DDA} \cite{gao2022back}: it performs input adaptation at test time via a diffusion model.
% % (3) \textbf{EATA} \cite{niu2022efficient}: it develops a sample-efficient entropy minimization strategy to avoid overfitting.
% (3) \textbf{MEMO} \cite{zhang2022memo}: it optimizes the entropy of the averaged prediction over multiple random augmentations of the input sample.
% (4) \textbf{TENT}~\cite{wang2020tent}: it fine-tunes scale and bias parameters of the batch normalization layers using an entropy minimization loss during inference.
% (5) \textbf{SAR} \cite{niutowards}: it encourages the model to lie in a flat minimum of the entropy loss surface with a reliable sampling strategy.

\begin{table*}[!htbp]
\begin{center}
\caption{Classification Accuracy (\%) for each corruption in \textbf{ImageNet-C} under \textbf{Imbalanced label shifts} at the highest severity.}
\label{table: imbalanced}
\resizebox{\linewidth}{!}
{
\begin{tabular}{l|ccccccccccccccc|c}
\toprule
Method & gaus & shot & impul & defcs & gls & mtn & zm & snw & frst & fg & brt & cnt & els & px & jpg & Avg.\\			
\midrule
Source  & 46.9 & 47.7 & 47.0 & 42.8 & 34.2 & 50.7 & 44.8 & 56.9 & 52.6 & 56.5 & 76.1 & 31.9 & 46.7 & 65.5 & 66.1 & 51.1\\
% EATA \cite{niu2022efficient} &  \\
DDA~\cite{gao2022back} &  52.6  & 54.0  & 52.2  & 33.7  & 40.8  & 33.6  & 30.2  & 29.8  & 35.0  & 5.0  & 48.8  & 2.7  & 50.2  & 60.2  & 58.9  & 39.2  \\
MEMO~\cite{zhang2022memo} & 58.1 & 59.1 & 58.5 	& 51.6 & 41.2 & 57.1 & 52.4 & 64.1 & 59.0 & 62.7 & \textbf{80.3} & 44.6 & 52.8 & 72.2 & 72.1 & 59.1 \\
TENT~\cite{wang2020tent}  & 58.5 & 59.9 & 59.9 & 58.6 & 57.2 & 62.5 & 59.3 & 67.0 & 28.9 & 71.0 & 79.3 & 62.9 & 65.5 & 73.8 & 71.9 & 62.4 \\
SAR~\cite{niutowards}  & \textbf{59.0} & 60.2 & 60.1 & \textbf{59.0} & 57.6 & 62.7 & 59.7 & 67.5 & 66.2 & 70.6 & 79.4 & 63.1 & 66.3 & 73.7 & 71.9 & 65.1 \\
\rowcolor{gray!20}
{\textbf{Ours}} & 58.8 & \textbf{60.5} & \textbf{60.2} & 58.9 & \textbf{58.6} & \textbf{63.6} & \textbf{62.6} & \textbf{69.1} & \textbf{68.3} & \textbf{72.8} & 79.5 & \textbf{63.9} & \textbf{69.1} & \textbf{74.3} & \textbf{72.5} & \textbf{66.2} \\
\rowcolor{gray!20}
& ${\pm0.1}$ & ${\pm0.1}$ & ${\pm0.2}$ & ${\pm0.1}$ & ${\pm0.3}$ & ${\pm0.1}$ & ${\pm0.5}$ & ${\pm0.2}$ & ${\pm0.3}$ & ${\pm0.2}$ & ${\pm0.1}$ & ${\pm0.8}$ & ${\pm0.4}$ & ${\pm0.4}$ & ${\pm0.1}$ & ${\pm0.1}$ \\
\bottomrule
\end{tabular}
}
% \vspace{-10pt}
\end{center}
\end{table*}

\begin{table*}[!htbp]
\begin{center}
\caption{Classification Accuracy (\%) for each corruption in \textbf{ImageNet-C} under \textbf{Batch size = 1} at the highest severity.}
\label{table: bs1}
\resizebox{\linewidth}{!}
{
\begin{tabular}{l|ccccccccccccccc|c}
\toprule
Method & gaus & shot & impul & defcs & gls & mtn & zm & snw & frst & fg & brt & cnt & els & px & jpg & Avg. \\
\midrule
Source  & 46.9 & 47.6 & 46.9 & 42.7 & 34.2 & 50.5 & 44.7 & 56.9 & 52.6 & 56.5 & 76.1 & 31.8 & 46.7 & 65.5 & 66.0 & 51.0  \\
DDA~\cite{gao2022back} & 52.5 & 54.0 & 52.1 & 33.8 & 40.6 & 33.3 & 30.2 & 29.7 & 35.0 & 5.0 & 48.6 & 2.7 & 50.0 & 60.0 & 58.8 & 39.1 \\
MEMO~\cite{zhang2022memo} & 58.1 & 59.1 & 58.5 	& 51.6 & 41.2 & 57.1 & 52.4 & 64.1 & 59.0 & 62.7 & \textbf{80.3} & 44.6 & 52.8 & 72.2 & 72.1 & 59.1 \\
TENT~\cite{wang2020tent}  & 58.6 & 60.1 & 60.0 & 59.0 & 57.4 & 62.7 & 59.7 & 67.3 & 45.5 & 71.4 & 79.2 & 63.9 & 66.1 & 73.9 & 71.9 & 63.8 \\
SAR~\cite{niutowards}  & 59.1 & 60.2 & 60.1 & 58.5 & 55.9 & 62.4 & 59.2 & 67.5 & 66.0 & 70.2 & 78.8 & 62.7 & 65.6 & 73.9 & 71.9 & 64.8 \\
\rowcolor{gray!20}
\textbf{Ours}  & \textbf{59.5}   & \textbf{61.0}   & \textbf{60.7}   & \textbf{59.2}   & \textbf{59.1}   & \textbf{63.8}   & \textbf{62.0}   & \textbf{69.6}   & \textbf{68.5}   & \textbf{73.5}   & 78.8  & \textbf{64.7}   & \textbf{68.8}   & \textbf{74.2}   & \textbf{72.4}   &  \textbf{66.4}   \\
\rowcolor{gray!20}
& ${\pm0.1}$ & ${\pm0.1}$ & ${\pm0.1}$ & ${\pm0.1}$ & ${\pm0.1}$ & ${\pm0.1}$ & ${\pm0.2}$ & ${\pm0.4}$ & ${\pm0.3}$ & ${\pm0.3}$ & ${\pm0.2}$ & ${\pm0.4}$ & ${\pm0.1}$ & ${\pm0.5}$ & ${\pm0.3}$ & ${\pm0.0}$ \\
\bottomrule
\end{tabular}
}
% \vspace{-10pt}
\end{center}
\end{table*}

\begin{table*}[!htbp]
\begin{center}
\caption{Classification Accuracy (\%) for test-time adaptation in \textbf{Office-Home} dataset.}
\label{table:officehome}
\resizebox{\linewidth}{!}
{
\begin{tabular}{l|cccccccccccc|c}
\toprule
Method & {A$\rightarrow$C} & {A$\rightarrow$P} & {A$\rightarrow$R} & {C$\rightarrow$A} & {C$\rightarrow$P} & {C$\rightarrow$R} & {P$\rightarrow$A} & {P$\rightarrow$C} & {P$\rightarrow$R} & {R$\rightarrow$A} & {R$\rightarrow$C} & {R$\rightarrow$P} & {Avg.} \\
\midrule
Source & 63.4 & 81.9 & 86.3 & 76.2 & 80.6 & 83.8 & 75.0 & 57.9 & 87.2 & 78.7 & 61.0 & 88.0 & 76.7 \\
TENT \cite{wang2020tent} & 69.1 & 81.8 & 86.5 & 76.5 & 81.9 & 83.2 & \textbf{76.8} & 65.0 & 86.7 & \textbf{81.1} & \textbf{69.7} & \textbf{88.2} & 78.9 \\
SAR~\cite{niutowards} & 67.3 & 80.7 & 85.6 & 77.5 & 79.8 & 84.1 & 74.7 & 60.3 & 87.6 & 78.9 & 63.1 & 87.7 & 77.3 \\
\rowcolor{gray!20}
\textbf{Ours} & \textbf{69.2} & \textbf{82.6} & \textbf{87.2} & \textbf{78.4} & \textbf{83.6} & \textbf{85.2} & \textbf{76.8} & \textbf{65.3} & \textbf{87.9} & 80.2 & 67.0 & 88.1 & \textbf{79.3} \\
\rowcolor{gray!20}
& ${\pm0.1}$ & ${\pm0.1}$ & ${\pm0.1}$ & ${\pm0.1}$ & ${\pm0.1}$ & ${\pm0.7}$ & ${\pm0.0}$ & ${\pm0.1}$ & ${\pm0.0}$ & ${\pm0.1}$ & ${\pm0.1}$ & ${\pm0.2}$ & ${\pm0.1}$ \\
\bottomrule
\end{tabular}
}
% \vspace{-10pt}
\end{center}
\end{table*}

% \footnote{SAR: {https://github.com/mr-eggplant/SAR}}
\subsection{Implementation Details} 
Following the official implementations of SAR, the pre-trained model weights are obtained from the \textit{timm} repository \cite{rw2019timm}. Specifically, for the Office-Home dataset, we fine-tune the ViT-B/16 model by replacing the original classifier head with a new classifier head. To ensure fair performance comparisons, all methods within each experimental condition share identical architecture and pre-trained model parameters. The domain-conditioner generator network is a fully connect layer with input dimension \textit{d} and output dimension \textit{3d}. The additional parameter number is about $1.7$ M for ViT-B/16.
The batch size is set to 64 for all experiments, except for the condition \textit{Batch size = 1}.
Our reported experimental results are the mean and standard deviation values obtained from three runs, each with random seeds chosen from the set \{2021, 2022, 2023\}. It should be noted that we use the matched normalization setting for the pre-trained \textit{timm} model (mean = [$0.5, 0.5, 0.5$], std = [$0.5, 0.5, 0.5$]), which is different from the code of the original SAR paper \cite{niutowards}. All models are tested on a single NVIDIA RTX3090 GPU. The source code will be released.
% More implementation details are provided in the Supplementary Materials.

\begin{figure*}[!ht]
\setlength{\abovecaptionskip}{0.1cm}
\setlength{\belowcaptionskip}{-0.1cm}
    \centering
    \includegraphics[width=\linewidth]{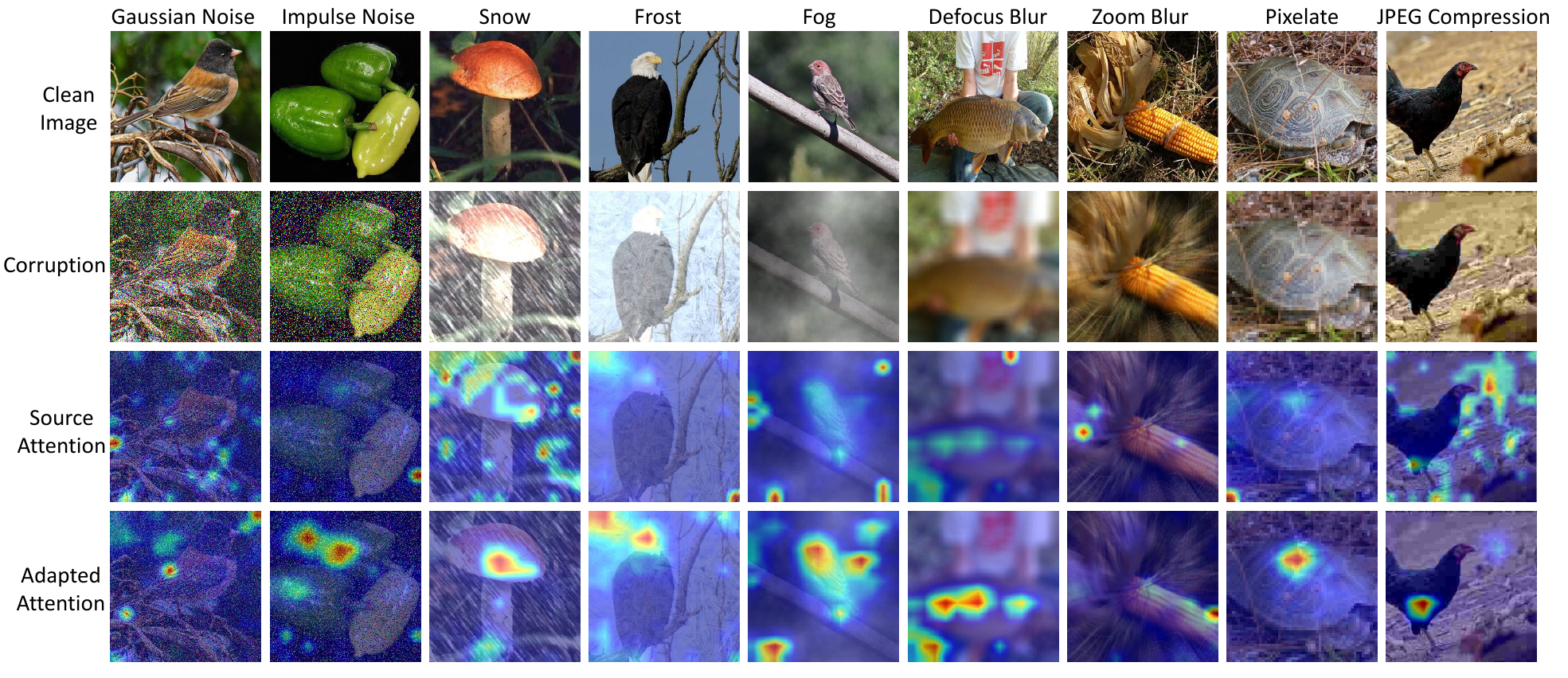}
    \caption{Representative examples of Attention Rollout \cite{abnar2020quantifying} for the source model ($3$-rd row) and our adapted model (last row).}
    \label{fig:attention}
\end{figure*}

\begin{table}[!ht]
    \centering
    \caption{Classification Accuracy (\%)  in \textbf{ImageNet-R} under \textbf{Normal} and \textbf{Batch size=1} settings.}
    \label{tab: imagenet-r}
    %\resizebox{0.7\linewidth}{!}{
    \begin{tabular}{l|cc}
    \toprule
        Method &  Normal & Batch size = 1\\
    \midrule
        Source &  57.2 &  57.2 \\
        % MEMO~\cite{zhang2022memo} & \\
        % EATA \cite{niu2022efficient} &  \\
        TENT \cite{wang2020tent}& 61.3 & 61.5\\
        SAR \cite{niutowards}& 62.0 & 61.8 \\
        \rowcolor{gray!20}
        \textbf{Ours} & \textbf{64.5} & \textbf{65.0} \\
        \rowcolor{gray!20}
         & ${\pm0.2}$ & ${\pm0.4}$ \\
    \bottomrule
    \end{tabular}   %}
% \vspace{-10pt}
\end{table}

\begin{table}[!ht]
    \centering
    \caption{Classification Accuracy (\%)  in \textbf{VisDA-2021} under \textbf{Normal} and \textbf{Batch size=1} settings.}
    \label{table:visda2021}
    % \resizebox{0.7\linewidth}{!}
    {
    \begin{tabular}{l|cc}
    \toprule
        Method &  Normal & Batch size = 1\\
    \midrule
        Source &  57.7 &  57.7 \\
        % MEMO~\cite{zhang2022memo} & \\
        % EATA \cite{niu2022efficient} &  \\
        TENT \cite{wang2020tent}& 60.1 & 60.1\\
        SAR \cite{niutowards}& 60.1 & 60.9 \\
        \rowcolor{gray!20}
        \textbf{Ours} & \textbf{62.2} & \textbf{62.7} \\
        \rowcolor{gray!20}
         & ${\pm0.2}$ & ${\pm0.5}$ \\
    \bottomrule
    \end{tabular}   
    }
% \vspace{-10pt}
\end{table}

\subsection{Performance Results} 
We evaluate the performance of our DCT method under three different test conditions on the ImageNet-C dataset. We report the official or reproduced top-1 accuracy using the official codes for all methods under comparison. Specifically,
% As discussed in the implementation details, all methods under comparison use the same transformer encoder. 
we first evaluate our approach under the \textbf{Normal} i.i.d assumption as shown in Table \ref{table: normal}. The results indicate that our approach achieves superior performance across the majority of the 15 corruption scenarios considered.
On average, our method outperforms the second-best method by 2.7\%. 
Subsequently, we evaluate our approach under the \textbf{Imbalanced label shifts} test condition as shown in Table \ref{table: imbalanced}. We can see that our method successfully enhances the mean classification accuracy across most of the corruption types, improving by 1.1\% on average. 
Furthermore, We evaluate our approach under the \textbf{Batch size = 1} test condition as shown in Table \ref{table: bs1}.
Our method improves the average classification accuracy of all 15 corruption types by 1.6\%,
demonstrating its adaptability with small batch sizes.

Additionally, we have extended our experimental analysis to include the \textbf{ImageNet-R} and \textbf{VisDA-2021} datasets. For ImageNet-R, we use the same pre-trained ViT-B/16 backbone and set the output size to 200 following the procedure in \cite{hendrycks2021many}. From  Table \ref{tab: imagenet-r} and Table \ref{table:visda2021}, we can see that our approach outperforms the previous state-of-the-art methods in both experimental settings, namely Normal and Batch size = 1. 
We extend our experimentation to \textbf{Office-Home} dataset. The results are presented in Table \ref{table:officehome}. The proposed DCT method outperforms the SAR method by 2.0\%. This further underscores the efficacy of our proposed DCT method across a diverse range of datasets.

In summary, our comprehensive results demonstrate the effectiveness and robustness of our proposed DCT approach in handling complex test conditions and outperforming state-of-the-art TTA methods across multiple evaluation metrics.
More experimental results are provided in the Supplementary Materials.

\subsection{Visualization and Discussion}

\begin{table}[!ht]
    \centering
    \caption{Ablation study under \textbf{Normal} at the highest severity. DC-generator represents the domain conditioner generator.}
\label{table: ablation}
% \resizebox{\linewidth}{!}
{
\begin{tabular}{l|cc}
\toprule
Method  & Avg. \\	
\midrule
% Source & 29.9 & 29.9\\
Baseline  Method & 63.6 \\
\quad + Domain-conditioner w/o DC-generator  & 63.9\\
\rowcolor{gray!20}
\textbf{Our DCT Method}  & \textbf{66.3}\\
\bottomrule
\end{tabular}
}
% % \end{minipage}
% \vspace{-5pt}
\end{table}

To explore the explainability of the domain-conditioned transformer, we visualize the attention map by Attention Rollout \cite{abnar2020quantifying} following ViT \cite{dosovitskiyimage}. As shown in Figure \ref{fig:attention}, given the corruption image, we can see that the adapted transformer attention focuses more on the object than the source model. This  demonstrates that our Domain-Conditioned Transformer (DCT) method significantly enhances the attention in the target domain.

Additionally, we performed an ablation study on the domain-conditioner generator in ImageNet-C dataset and ViT-B/16 backbone as shown in Table \ref{table: ablation}. When solely integrating learnable domain-conditioners into query, key, and value without the domain-conditioner generator's conditional generation based on the class token, the average accuracy improved by $0.3$\%. In contrast, when adapting the domain-conditioner generator to generate domain conditioners conditioned by the class token, we observed a substantial improvement of $2.7$\%. It demonstrates the significant contribution of the class token conditioned domain-conditioner generator in enhancing the model's performance.

\begin{figure}[!ht]
% \vspace{-10pt}
\setlength{\abovecaptionskip}{0.1cm}
    \centering
    \includegraphics[width=\linewidth]{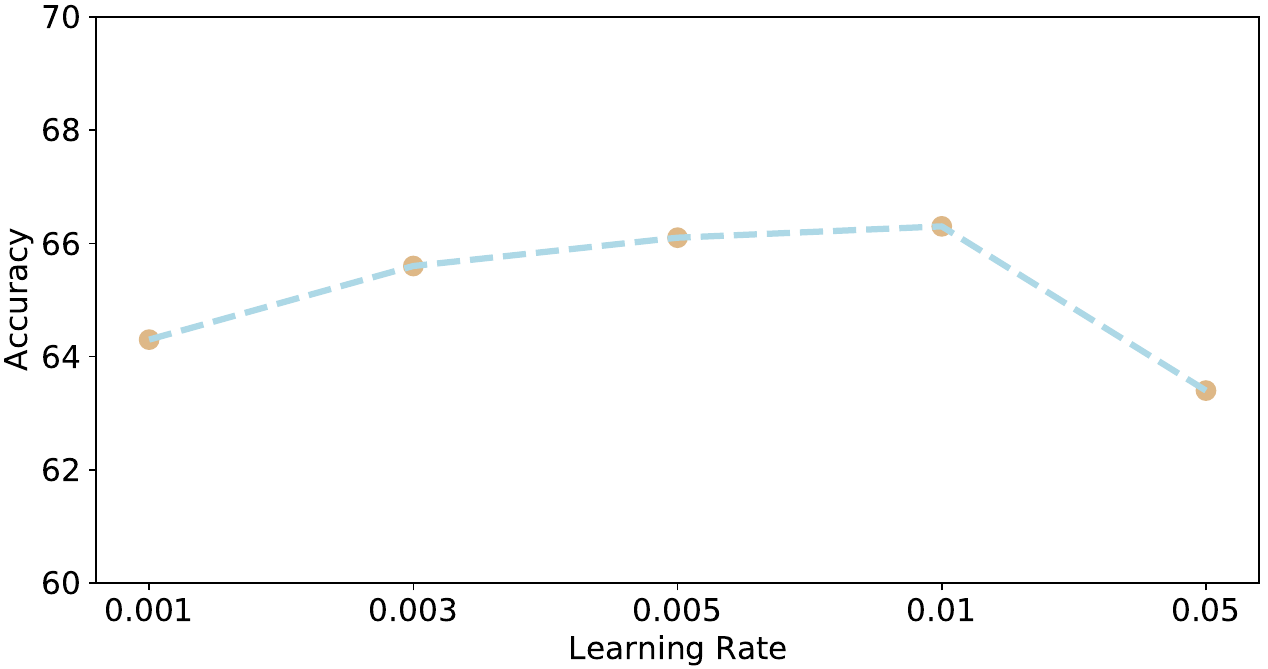}
    \caption{Sensitivity analyses for the learning rates of the domain-conditioner generator.}
    \label{fig:sensitivity}
% \vspace{-10pt}
\end{figure}

We conduct parameter sensitivity analysis on the learning rates for the domain-conditioner generator with the Normal setting in ImageNet-C with ViT-B/16.  As shown in Figure \ref{fig:sensitivity}, we can see that the performance is best when the learning rate of the domain-conditioner generator is set to $0.01$.

\section{Conclusion}
\label{sec:conclusion}
Fully test-time adaptation is a challenging problem in computer vision, particularly in the presence of complex corruptions and shifts in the test data distribution. 
In this work, we have tackled the critical challenge of adapting transformer-based models to new domains, focusing on the significant deviation in self-attention profiles encountered in the target domain compared to the source domain.
Specifically, we have introduced three domain-conditioning vectors, called domain conditioners into the self-attention module.
By integrating these domain conditioners into the query, key, and value components of the self-attention module, we have effectively mitigated the impact of domain shift observed during inference. The dynamic generation of these domain conditioners at each transformer network layer, derived from the class token, allowed for a gradual removal of domain shift effects, thereby enabling the recovery of the original self-attention profile in the target domain.
Our experimental results demonstrated that our proposed DCT method is able to significantly improve fully test-time adaptation performance.

\section*{Acknowledgements}
This research is supported by the National Natural Science Foundation of China (No. 62331014) and Project 2021JC02X103.

\bibliographystyle{ACM-Reference-Format}
\bibliography{sample-base}

% \appendix









\end{document}

\title{Supplementary Materials: Domain-Conditioned Transformer for Fully Test-time Adaptation}

% \affiliation{%

% \author{Anonymous Authors}
\author{Yushun Tang}
\affiliation{%
  \institution{Southern University of Science and Technology}
  % \streetaddress{1 Th{\o}rv{\"a}ld Circle}
  \city{Shenzhen}
  \country{China}}
\email{tangys2022@mail.sustech.edu.cn}

\author{Shuoshuo Chen}
\affiliation{%
  \institution{Southern University of Science and Technology}
  % \streetaddress{1 Th{\o}rv{\"a}ld Circle}
  \city{Shenzhen}
  \country{China}}
\email{chenss2021@mail.sustech.edu.cn}

\author{Jiyuan Jia}
\affiliation{%
  \institution{Southern University of Science and Technology}
  % \streetaddress{1 Th{\o}rv{\"a}ld Circle}
  \city{Shenzhen}
  \country{China}}
\email{jiajy2018@mail.sustech.edu.cn}

\author{Yi Zhang}
\affiliation{%
  \institution{Southern University of Science and Technology}
  % \streetaddress{1 Th{\o}rv{\"a}ld Circle}
  \city{Shenzhen}
  \country{China}}
\email{zhangyi2021@mail.sustech.edu.cn}

\author{Zhihai He}
\authornote{Corresponding author.}
\affiliation{%
  \institution{Southern University of Science and Technology, Pengcheng Laboratory}
  % \streetaddress{1 Th{\o}rv{\"a}ld Circle}
  \city{Shenzhen}
  \country{China}}
\email{hezh@sustech.edu.cn}

\maketitle

\appendix
In this Supplementary Materials, we provide more details and experimental results for further understanding of the proposed Domain-Conditioned Transformer method.

\section{Running Time Comparisons}

We provide a comprehensive comparison of running time costs in Table \ref{table: cost}. Our proposed DCT model introduces a slightly higher computational overhead, approximately 13\% slower than the SAR method, but much faster than MEMO and DDA.
This extra overhead is mainly caused by the learning of the domain conditioner generator. In the original SAR method, it only updates the layer normalization parameters. 

\begin{table}[!htbp]
    \centering
        \caption{Testing time cost comparison for Gaussian corruption of \textbf{ImageNet-C} under the single GPU NVIDIA RTX 3090.}
    \label{table: cost}
    % \resizebox{0.2\linewidth}{!}
    {
    \begin{tabular}{l|l}
    \toprule
        Method & Time Cost \\
    \midrule
          TENT & $\sim$ 5 min \\
          SAR & $\sim$ 7 min \\
          MEMO & $\sim$ 14 hours \\
          DDA & $\sim$ 5 days \\
          \textbf{Ours} & $\sim$ 8 min \\
    \bottomrule
    \end{tabular}}
\vspace{-15pt}
\end{table}

% \section{Additional Domain-Conditioner Visualization}
% Figure \ref{fig:qkv_layer0} shows the t-SNE plot of the additional domain conditioners $[C^l_q, C^l_k, C^l_v]$ in layer 1 for samples from three different target domains, plotted with three different colors.   
% We can see that samples from the same domain aggregate together. This further suggests that the domain conditioners $[C^l_q, C^l_k, C^l_v]$ generated by the learned network $\Phi^l$ are able to capture the domain characteristics. 

% \begin{figure}[!ht]
% \vspace{-10pt}
% \setlength{\abovecaptionskip}{0.1cm}
% % \setlength{\belowcaptionskip}{-0.2cm}
%     \centering
%     \includegraphics[width=0.9\linewidth]{figs/qkv_layer0.pdf}
%     \caption{Visualization of the domain conditioners from the first transformer layer.}
%     \label{fig:qkv_layer0}
% \vspace{-10pt}
% \end{figure}

\section{Additional Results on ImageNet-C with Severity Level 3}

\begin{table*}[!htbp]
\begin{center}
\caption{Classification Accuracy (\%) for each corruption in \textbf{ImageNet-C} under \textbf{Normal} at the severity Level 3. The best result is shown in \textbf{bold}.}
\label{table: normal_3}
\resizebox{\linewidth}{!}
{
\begin{tabular}{l|ccccccccccccccc|c}
\toprule
Method & gaus & shot & impul & defcs & gls & mtn & zm & snw & frst & fg & brt & cnt & els & px & jpg & Avg.\\	
\midrule
Source & 72.1 & 71.5 & 71.3 & 62.3 & 51.1 & 69.1 & 59.1 & 69.0 & 60.0 & 70.1 & 80.1 & 74.0 & 75.1 & 77.6 & 75.2 & 69.2 \\
% MEMO & \\
DDA & 62.6  & 63.1  & 62.3  & 50.2  & 54.2  & 50.9  & 43.3  & 41.0  & 41.9  & 14.8  & 60.6  & 26.0  & 62.2  & 62.6  & 62.8  & 50.6 \\
TENT & 74.3 & 73.9 & 73.6 & 70.8 & 66.6 & 73.7 & 66.9 & 73.2 & 68.7 & 76.0 & 81.6 & 78.9 & 78.5 & 79.7 & 77.1 & 74.2 \\
SAR & 74.3 & 73.9 & 73.7 & 70.9 & 66.5 & 73.8 & 66.9 & 73.1 & 68.7 & 75.8 & 81.8 & 78.9 & 78.5 & 79.8 & 77.1 & 74.2 \\ 
\rowcolor{gray!20}
\textbf{Ours} & \textbf{74.7} & \textbf{74.6} & \textbf{74.4} & \textbf{71.3} & \textbf{69.6} & \textbf{74.4} & \textbf{70.1} & \textbf{74.8} & \textbf{71.6} & \textbf{78.1} & \textbf{81.9} & \textbf{79.6} & \textbf{79.3} & \textbf{80.1} & \textbf{78.7} & \textbf{75.5} \\ 
\rowcolor{gray!20}
& ${\pm0.1}$ & ${\pm0.0}$ & ${\pm0.1}$ & ${\pm0.0}$ & ${\pm0.1}$ & ${\pm0.1}$ & ${\pm0.2}$ & ${\pm0.0}$ & ${\pm0.1}$ & ${\pm0.1}$ & ${\pm0.1}$ & ${\pm0.2}$ & ${\pm0.1}$ & ${\pm0.0}$ & ${\pm0.2}$ & ${\pm0.0}$ \\
\bottomrule
\end{tabular}
}
% \vspace{-10pt}
\end{center}
\end{table*}

\begin{table*}[!htbp]
\begin{center}
\caption{Classification Accuracy (\%) for each corruption in \textbf{ImageNet-C} under \textbf{Imbalanced label shifts} at the severity Level 3.}
\label{table: imbalanced_3}
\resizebox{\linewidth}{!}
{
\begin{tabular}{l|ccccccccccccccc|c}
\toprule
Method & gaus & shot & impul & defcs & gls & mtn & zm & snw & frst & fg & brt & cnt & els & px & jpg & Avg.\\			
\midrule
Source & 51.5 & 46.8 & 50.4 & 48.7 & 37.1 & 54.7 & 41.6 & 35.1 & 33.3 & 68.0 & 69.3 & 74.9 & 65.9 & 66.0 & 63.6 & 53.8 \\
% MEMO & \\
DDA & 62.6  & 63.1  & 62.3  & 50.2  & 54.2  & 50.9  & 43.3  & 41.0  & 41.9  & 14.8  & 60.6  & 26.0  & 62.2  & 62.6  & 62.8  & 50.6 \\
TENT & 74.6 & 74.3 & 74.0 & 71.4 & 68.4 & 74.6 & 68.3 & 73.7 & 69.8 & 77.0 & 81.9 & 79.2 & 79.2 & 80.0 & 78.1 & 75.0 \\
SAR & \textbf{74.7} & 74.4 & 74.2 & \textbf{71.5} & 68.7 & \textbf{74.8} & 68.6 & 74.0 & 70.2 & 77.0 & \textbf{82.1} & 79.4 & 79.3 & \textbf{80.2} & 78.2 & 75.1 \\ 
\rowcolor{gray!20}
\textbf{Ours} & 74.6 & \textbf{74.5} & \textbf{74.3} & \textbf{71.5} & \textbf{69.5} & 74.7 & \textbf{69.8} & \textbf{75.0} & \textbf{71.4} & \textbf{77.9} & 81.9 & \textbf{79.6} & 79.2 & \textbf{80.2} & \textbf{78.9} & \textbf{75.5}  \\ 
\rowcolor{gray!20}
& ${\pm0.2}$ & ${\pm0.1}$ & ${\pm0.1}$ & ${\pm0.1}$ & ${\pm0.2}$ & ${\pm0.0}$ & ${\pm0.2}$ & ${\pm0.1}$ & ${\pm0.1}$ & ${\pm0.3}$ & ${\pm0.1}$ & ${\pm0.1}$ & ${\pm0.2}$ & ${\pm0.1}$ & ${\pm0.1}$ & ${\pm0.1}$ \\
\bottomrule
\end{tabular}
}
% \vspace{-10pt}
\end{center}
\end{table*}

\begin{table*}[!htbp]
\begin{center}
\caption{Classification Accuracy (\%) for each corruption in \textbf{ImageNet-C} under \textbf{Batch size=1} at the severity Level 3.}
\label{table: bs1_3}
\resizebox{\linewidth}{!}
{
\begin{tabular}{l|ccccccccccccccc|c}
\toprule
Method & gaus & shot & impul & defcs & gls & mtn & zm & snw & frst & fg & brt & cnt & els & px & jpg & Avg.\\			
\midrule
Source & 51.6 & 46.9 & 50.5 & 48.7 & 37.2 & 54.7 & 41.6 & 35.1 & 33.5 & 67.8 & 69.3 & 74.8 & 65.8 & 66.0 & 63.7 & 53.8 \\
% MEMO & \\
DDA & 62.6  & 63.1  & 62.3  & 50.2  & 54.2  & 50.9  & 43.3  & 41.0  & 41.9  & 14.8  & 60.6  & 26.0  & 62.2  & 62.6  & 62.8  & 50.6 \\
TENT & 74.4 & 73.9 & 73.6 & 70.9 & 66.6 & 73.7 & 67.0 & 73.1 & 68.7 & 76.0 & 81.6 & 79.0 & 78.5 & 79.8 & 77.1 & 74.3  \\
SAR & \textbf{74.9} & \textbf{74.6} & 74.0 & 71.4 & 68.2 & \textbf{74.5} & 68.2 & \textbf{73.9} & 70.0 & 76.3 & \textbf{81.2} & 78.9& \textbf{78.4} & \textbf{79.3} & 77.1 & 74.7 \\ 
\rowcolor{gray!20}
\textbf{Ours} & 74.6 & 74.4 & \textbf{74.1} & \textbf{71.5} & \textbf{69.6} & 74.4 & \textbf{69.1} & 73.4 & \textbf{71.7} & \textbf{77.4} & 81.1 & \textbf{79.0} & 77.9 & 79.2 & \textbf{78.1} & \textbf{75.0} \\ 
\rowcolor{gray!20}
& ${\pm0.1}$ & ${\pm0.3}$ & ${\pm0.2}$ & ${\pm0.2}$ & ${\pm0.2}$ & ${\pm0.3}$ & ${\pm0.3}$ & ${\pm0.2}$ & ${\pm0.1}$ & ${\pm0.5}$ & ${\pm0.2}$ & ${\pm0.2}$ & ${\pm0.3}$ & ${\pm0.1}$ & ${\pm0.2}$ & ${\pm0.0}$ \\  
\bottomrule
\end{tabular}
}
% \vspace{-10pt}
\end{center}
\end{table*}

We provide additional performance comparison results for corruption severity Level 3 with the Normal, Imbalanced label shifts, and Batch size = 1 settings in Table \ref{table: normal_3}, \ref{table: imbalanced_3}, and \ref{table: bs1_3}, respectively. The results are consistent with those in the main paper for severity level 5. We can see that our DCT method outperforms existing methods in almost all 15 corruption types.

% \section{Additional Results on ImageNet-C with Mixed shifts}
% We conduct performance comparisons under the more complex \textbf{Mixed shifts} test condition. We report the results at both Level 5 and Level 3 corruptions. As shown in Table \ref{table: o_mixed}, our DCT algorithm only improves the accuracy by 0.2\%. This is because our algorithm aims to learn the domain-specific visual conditioning token. This learning becomes less effective when the target samples have mixed domains. 

% \begin{table}[!ht]
% % \begin{minipage}{0.4\textwidth}
% % \setlength{\abovecaptionskip}{0cm}
% % \setlength{\belowcaptionskip}{-0.2cm}
%     \centering
%     \caption{Top-1 Classification average Accuracy (\%) under \textbf{Mixed shifts} at Level 5 and Level 3.}\label{table: o_mixed}
%     % \resizebox{0.8\linewidth}{!}
%     {
%     \begin{tabular}{l|cc}
%     \toprule
%         Method & Level 5  & Level 3 \\
%     \midrule
%         Source & 51.0  & 69.2  \\
%         TENT    & 37.8    & 72.8    \\
%         % EATA \cite{niu2022efficient}  & 55.7  & 69.6   \\
%         SAR  & 61.3  & 73.3  \\
%         \rowcolor{gray!20}
%         \textbf{Ours} & \textbf{61.5}  &  \textbf{73.4} \\ 
%         \rowcolor{gray!20}
%         & ${\pm0.0}$ & ${\pm0.0}$\\
%     \bottomrule
%     \end{tabular}
%     }
% \end{table}

\section{Additional Results on ImageNet-C with ViT-L/16 Backbone}
We extend our experimentation to encompass a larger ViT-L/16 backbone, operating within the contexts of both Normal and Batch size = 1 settings. For the Normal setting, the batch size is set to 32 due to limited memory. The learning rate of the domain-conditioner generator is set 
to $0.001$ and $0.0001$ respectively. The results, as illustrated in Table \ref{tab: vit-l-normal} and \ref{tab: vit-l-bs1}, consistently showcase the superiority of our proposed DCT method over the baseline SAR in both configurations. This robust performance demonstrates the efficiency of our proposed DCT method across diverse transformer backbones.

\begin{table*}[!ht]
    \centering
    \caption{Classification Accuracy (\%)  in \textbf{ImageNet-C} with \textbf{ViT-L/16} under \textbf{Normal} at the highest severity (Level 5).}
    \label{tab: vit-l-normal}
    \resizebox{\linewidth}{!}{
\begin{tabular}{l|ccccccccccccccc|c}
\toprule
Method & gaus & shot & impul & defcs & gls & mtn & zm & snw & frst & fg & brt & cnt & els & px & jpg & Avg.\\
    \midrule
        Source & 62.1 & 61.4 & 62.3 & 52.7 & 45.1 & 60.6 & 55.1 & 66.2 & 62.4 & 62.5 & 80.2 & 39.8 & 56.2 & 74.3 & 72.7 & 60.9 \\
        % DDA & 55.73  & 57.21  & 55.46  & 44.11  & 49.54  & 42.02  & 38.50  & 34.96  & 39.55  & 7.97  & 54.68  & 5.95  & 54.80  & 65.53  & 63.25  & 44.62 \\
        TENT & 65.6 & 68.3 & 67.6 & 63.4 & 59.9 & 66.8 & 60.7 & 69.0 & 68.5 & 67.4 & 81.0 & 28.9 & 64.7 & 77.2 & 74.7 & 65.6 \\
        SAR & 66.0 & 66.6 & 66.2 & 61.3 & 55.1 & 66.1 & 58.3 & 68.4 & 65.7 & 66.3 & 81.0 & 26.8 & 63.7 & 74.5 & 73.6 & 64.0 \\
        \rowcolor{gray!20}
        \textbf{Ours} & \textbf{67.1} & \textbf{68.2} & \textbf{66.9} & \textbf{64.0} & \textbf{62.4} & \textbf{66.7} & \textbf{64.1} & \textbf{71.1} & \textbf{69.1} & \textbf{70.4} & \textbf{81.3} & \textbf{65.3} & \textbf{69.7} & \textbf{77.6} & \textbf{75.5} & \textbf{69.3} \\
        \rowcolor{gray!20}
        & ${\pm0.7}$ & ${\pm0.4}$ & ${\pm0.9}$ & ${\pm1.0}$ & ${\pm1.3}$ & ${\pm0.5}$ & ${\pm0.5}$ & ${\pm0.9}$ & ${\pm0.5}$ & ${\pm0.7}$ & ${\pm0.1}$ & ${\pm0.6}$ & ${\pm1.9}$ & ${\pm0.1}$ & ${\pm0.5}$ & ${\pm0.3}$ \\
    \bottomrule
    \end{tabular}   }
\end{table*}

\begin{table*}[!ht]
    \centering
    \caption{Classification Accuracy (\%)  in \textbf{ImageNet-C} with \textbf{ViT-L/16} under \textbf{Batch size = 1} at the highest severity (Level 5).}
    \label{tab: vit-l-bs1}
    \resizebox{\linewidth}{!}{
\begin{tabular}{l|ccccccccccccccc|c}
\toprule
Method & gaus & shot & impul & defcs & gls & mtn & zm & snw & frst & fg & brt & cnt & els & px & jpg & Avg.\\
    \midrule
        Source & 62.1 & 61.4 & 62.3 & 52.7 & 45.1 & 60.6 & 55.1 & 66.2 & 62.4 & 62.5 & 80.2 & 39.8 & 56.2 & 74.3 & 72.7 & 60.9 \\
        % DDA & 55.73  & 57.21  & 55.46  & 44.11  & 49.54  & 42.02  & 38.50  & 34.96  & 39.55  & 7.97  & 54.68  & 5.95  & 54.80  & 65.53  & 63.25  & 44.62 \\
        TENT & 55.2 & 67.8 & 67.8 & 43.5 & 59.6 & 66.9 & 62.8 & 69.9 & 67.4 & 68.3 & 81.4 & 31.4 & 64.7 & 77.4 & 73.0 & 63.8 \\
        SAR & \textbf{67.9} & 64.5 & 67.7 & 63.0 & 61.6 & 63.7 & 62.4 & 70.4 & 67.6 & 68.7 & 75.7 & 60.2 & 55.5 & 76.4 & 74.7 & 66.7 \\
        \rowcolor{gray!20}
        \textbf{Ours} & 67.5 & \textbf{69.2} & \textbf{67.9} & \textbf{64.6} & \textbf{64.5} & \textbf{68.5} & \textbf{65.7} & \textbf{71.0} & \textbf{68.8} & \textbf{71.6} & \textbf{80.6} & \textbf{65.0} & \textbf{71.2} & \textbf{76.5} & \textbf{75.3} & \textbf{69.9} \\
        \rowcolor{gray!20}
        & ${\pm0.3}$ & ${\pm0.2}$ & ${\pm0.0}$ & ${\pm0.6}$ & ${\pm0.9}$ & ${\pm0.1}$ & ${\pm1.6}$ & ${\pm0.5}$ & ${\pm0.7}$& ${\pm0.0}$ & ${\pm0.1}$ & ${\pm0.4}$ & ${\pm1.7}$ & ${\pm0.2}$ & ${\pm0.7}$ & ${\pm0.3}$ \\
    \bottomrule
    \end{tabular}   }
\end{table*}

% \bibliographystyle{ACM-Reference-Format}

% \appendix







